\pdfoutput=1

\documentclass[11pt]{article}

\usepackage[final]{acl}

\usepackage{times}
\usepackage{latexsym}
\usepackage{adjustbox}
\usepackage{caption}

\usepackage[T1]{fontenc}

\usepackage[utf8]{inputenc}

\usepackage{microtype}

\usepackage{inconsolata}

\usepackage{diagbox}
\usepackage{chngpage}
\usepackage{times}
\usepackage{url}
\usepackage{latexsym}
\usepackage{booktabs}
\usepackage{graphicx}

\usepackage{multirow}
\usepackage{amsmath}
\usepackage{interval}
\usepackage{amssymb}
\usepackage{float}

\usepackage{enumitem}
\usepackage[utf8]{inputenc}

\usepackage{xargs}
\usepackage{placeins}
\usepackage{listings}

\usepackage{graphicx}

\title{SciClaims: An End-to-End Generative System for Biomedical Claim Analysis}

\author{Raúl Ortega \\
  Language Technology Research Lab \\
  Expert.ai / Madrid, Spain\\
  \texttt{rortega@expert.ai} \\\And
  José Manuel Gómez-Pérez \\
  Language Technology Research Lab \\
  Expert.ai / Madrid, Spain \\
  \texttt{jmgomez@expert.ai} \\}

\begin{document}
\maketitle
\begin{abstract}

We present SciClaims, an interactive web-based system for end-to-end scientific claim analysis in the biomedical domain. Designed for high-stakes use cases such as systematic literature reviews and patent validation, SciClaims extracts claims from text, retrieves relevant evidence from PubMed, and verifies their veracity. The system features a user-friendly interface where users can input scientific text and view extracted claims, predictions, supporting or refuting evidence, and justifications in natural language. Unlike prior approaches, SciClaims seamlessly integrates the entire scientific claim analysis process using a single large language model, without requiring additional fine-tuning. SciClaims is optimized to run efficiently on a single GPU and is publicly available for live interaction\footnote{\url{https://labdemos.expertcustomers.ai/health\_claims} (user: guest / password: acldemos2025)}. 
\end{abstract}

\section{Introduction}
Systematic Literature Review (SLR) plays a critical role in biomedical research and the pharmaceutical industry, supporting clinical decisions, regulatory submissions, and R\&D pipelines. A central task in SLR is the validation of scientific claims, ensuring that assertions made in scientific texts are supported by prior peer-reviewed research. However, this task is labor-intensive, prone to errors, and increasingly difficult to scale as the number of scientific publications grows.

We present SciClaims, a fully automated system that addresses this challenge by providing an end-to-end scientific claim analysis pipeline in an interactive, user-friendly interface. The system extracts factual claims from scientific texts, retrieves evidence from a curated biomedical corpus, and verifies the validity of each claim using large language models (LLMs). To improve transparency, SciClaims also offers natural language rationales and highlights key supporting or refuting evidence.

SciClaims is optimized for real-world deployment and operates efficiently on a single GPU, supports high-throughput processing, and handles documents of up to 10,000 characters. It is accessible through a web-based interface, allowing users to analyze both preloaded and custom input texts.

In this demonstration, we showcase SciClaims as a useful tool to enable users to validate scientific claims in real time, facilitating trustworthy knowledge discovery in high-stakes domains like biomedicine and pharmaceuticals. A how-to video\footnote{\url{www.youtube.com/watch?v=jyms_Ey0YSQ}} and all the code\footnote{ \url{www.github.com/expertailab/sciclaims-backend} and \url{www.github.com/expertailab/sciclaims-frontend}} used in this project have been published and made publicly available.

\begin{figure*}[ht]
    \centering
    \includegraphics[width=0.979\textwidth]{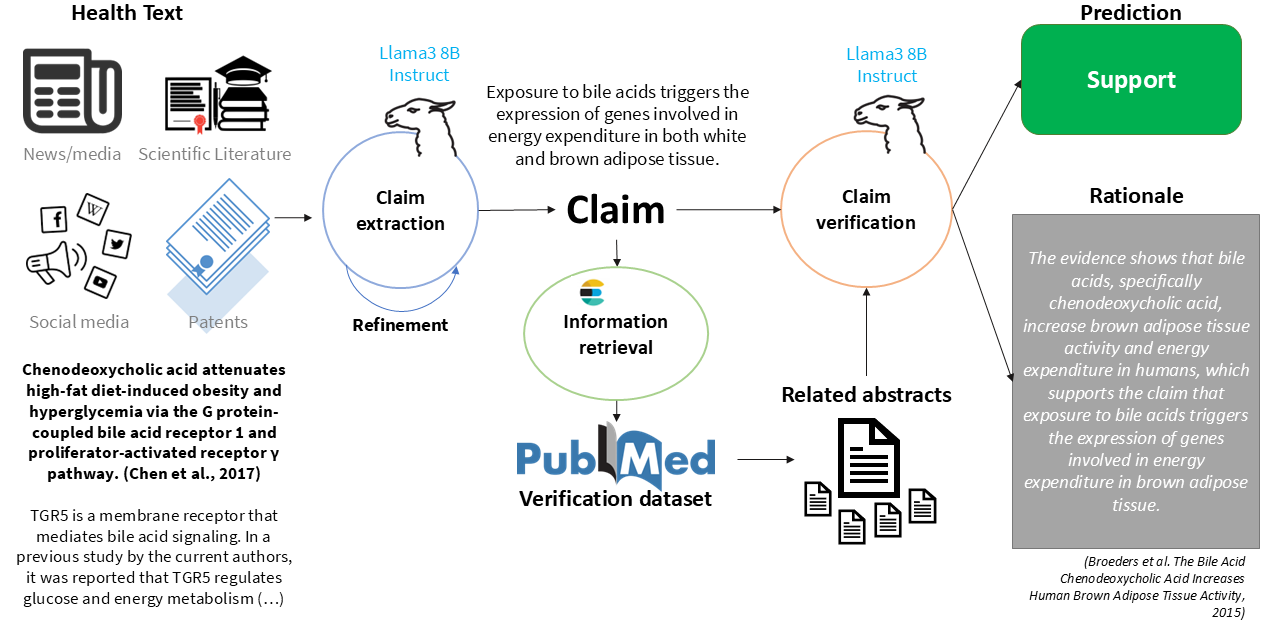}
    \caption{System Architecture.}
   \label{fig:arch}
\end{figure*}

\section{Related Work}
The task of analyzing  scientific claims from real-world texts based on background knowledge consists of three primary components: claim extraction, evidence retrieval from a document corpus, and verifying or fact-checking the claims against the evidence~\cite{eldifrawi-etal-2024-automated, vladika2023scientificfactcheckingsurveyresources}. However, solving this pipeline end-to-end across all three stages remains an open challenge.

Several studies have addressed the challenge of extracting claims from scientific texts. Some frameworks based on zero-shot models~\cite{pan-etal-2021-zero, wright-etal-2022-generating} have achieved promising results, primarily focusing on generating claim datasets from raw texts to train fact-checking models in specific domains. However, these methods rely on multi-stage NLP pipelines that are prone to failure and tend to produce a large number of claims per document, making their integration into an end-to-end system difficult. In contrast, recent approaches based on LLMs extract atomic factual units~\cite{min-etal-2023-factscore, chern2023factoolfactualitydetectiongenerative}, which serve as concise, interpretable summaries of the source texts.

For the evidence retrieval phase, dense passage retrieval methods, such as ColBERT \cite{khattab2020colbertefficienteffectivepassage}, have emerged due to their ability to retrieve highly relevant documents from large corpora with great precision. However, these methods are computationally expensive and therefore less practical for real-time, lightweight applications. Thus, simpler approaches such as BM-25~\cite{Robertson2009ThePR} and tools like Elasticsearch, known for their balance between retrieval quality and computational efficiency, are still preferred for deployment in production environments.

The release of several claim verification datasets has spurred the development of numerous models aimed at addressing this challenge. VERT5ERINI~\cite{pradeep-etal-2021-scientific}, PARAGRAPHJOINT~\cite{li2021paragraphlevelmultitasklearningmodel}, and MultiVerS~\cite{wadden2022multiversimprovingscientificclaim} have achieved promising results on scientific benchmarks like SciFact~\cite{wadden-etal-2020-fact} and CLIMATE-FEVER~\cite{diggelmann2021climatefeverdatasetverificationrealworld}. However, these approaches still leave room for improvement and typically rely on large, labeled datasets, which limits their scalability across domains. In parallel, recent advances in LLMs have led to increased interest in verifying automatically generated content. Frameworks such as FactScore~\cite{min-etal-2023-factscore}, FacTool~\cite{chern2023factoolfactualitydetectiongenerative}, and LLM Oasis~\cite{scirè2024truthmirageendtoendfactuality} focus on assessing the factuality of LLM-generated text rather than existing, human-written passages. Nonetheless, their techniques, such as extracting atomic knowledge units and using zero-shot LLMs for claim extraction and verification, can be extended to real-world scientific texts, offering a promising direction for scalable, data-efficient claim analysis.

Although individual components have seen substantial progress, an integrated system capable of seamlessly connecting these steps remains an open problem. Many existing systems, such as FactDetect \cite{jafari2024robustclaimverificationfact}, CliVER \cite{10.1093/jamiaopen/ooae021}, and more recently \cite{liang2025, vladika-etal-2025-step}, focus on claims that are already identified, neglecting the crucial first step of extracting relevant claims from raw, real-world texts. As a result, these systems are limited to pre-identified claims rather than addressing the full pipeline of claim extraction, retrieval, and verification.

Given the complexity of this problem, our approach aims to optimize each stage of the pipeline to ensure both efficiency and accuracy in real-world biomedical claim analysis. In this work, we tap on the strengths of modern LLMs to address previous limitations, specifically in the claim extraction and verification stages, and integrate our results as a robust online system.

\section{System Description}
\label{sec:system}

\begin{figure*}[ht]
    \centering
    \includegraphics[width=0.979\textwidth]{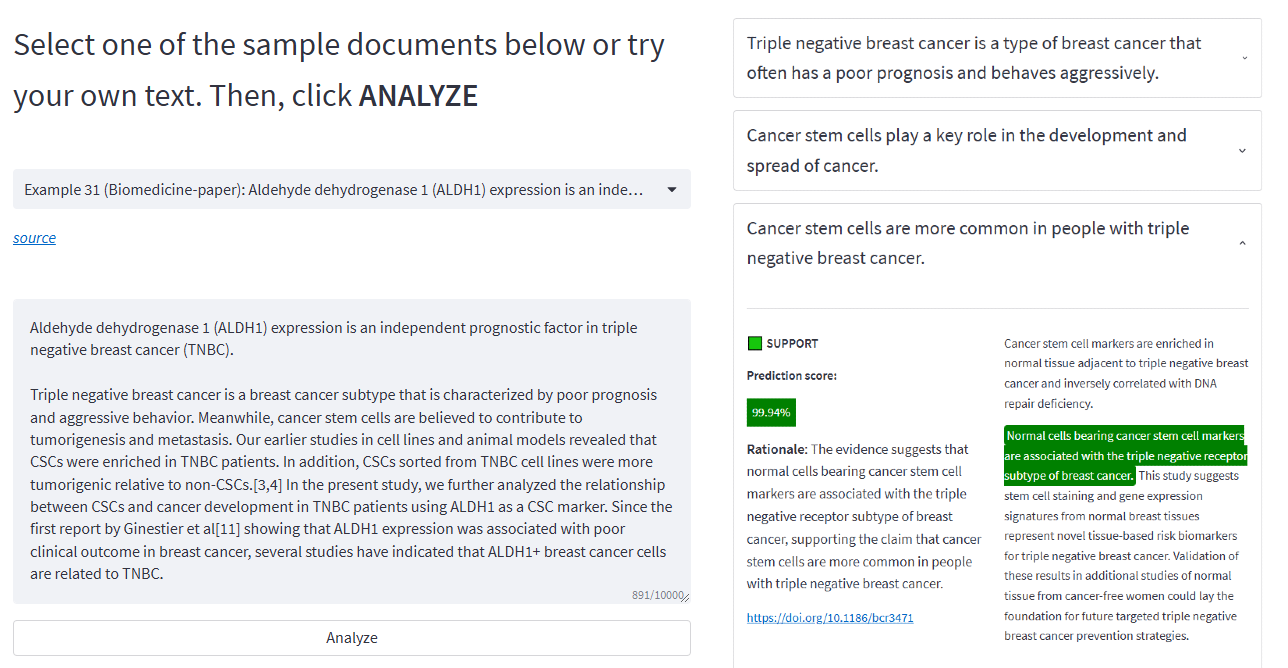}
    \caption{Screenshot of the SciClaims Demo.}
   \label{fig:screenshot}
\end{figure*}

SciClaims is an interactive, end-to-end system for scientific claim analysis, comprising three main components: claim extraction, evidence retrieval, and claim verification. It is optimized to run efficiently on a single 24GB VRAM GPU, enabling real-time performance via a web-based interface.

The system architecture (Figure \ref{fig:arch}) is built around a Llama3 8B Instruct model \cite{grattafiori2024llama3herdmodels} and an Elasticsearch-based retrieval engine. It processes biomedical or scientific text input and outputs a list of extracted claims, their verification status, relevant evidence, and a natural language rationale. The model is set up using vLLM \cite{kwon2023efficientmemorymanagementlarge}, which enables high-throughput processing in inference. Next, we present the main building blocks of SciClaims.

\textbf{Claim extraction}: This first module extracts potential claims from the source text. A claim is characterized by a specific set of properties detailed in section~\ref{sec:exp}. The Claim Extraction module calls the Llama3 8B Instruct model twice: first, to generate an initial list of claims, and second, to refine and filter them, improving the quality of the resulting claims. All the prompts used in our pipeline are provided in Appendix \ref{sec:prompts}. Based on our evaluation, we made adjustments to the initial prompts to improve claim extraction performance. Further details on this refinement can be found in section~\ref{sec:exp}.

\textbf{Document retrieval}: The second module retrieves potentially relevant documents from the verification dataset, using the claim as a query to the Elasticsearch index. The verification dataset contains 4.7 million abstracts from PubMed (2000–2022) that were curated using the Semantic Scholar's \textit{Highly Influential Citations} metric~\cite{valenzuela2015identifying}, ensuring that each article is backed by at least three highly influential citations. This selection criterion helps prioritize documents that have been extensively referenced in the academic community, enhancing the quality and relevance of retrieved information for verification. 
We chose not to filter documents based on Elasticsearch's scoring mechanism to maximize recall. Instead, the subsequent verification module is responsible for discarding irrelevant documents.

\textbf{Claim verification}: The final module performs fact-checking by making another call to the LLM, providing the claim along with the retrieved related documents. The model assigns one of the following three labels: 1) \textit{SUPPORT} if the claim is verified by the document, 2) \textit{REFUTE} if the claim is refuted by the document, or 3) \textit{NEI} (not enough information) if the document lacks sufficient evidence or is not relevant to the claim. To improve transparency and interpretability, we also request the model to provide a rationale for its decision, including identifying the most relevant sentence(s) that support its conclusion.

\section{User interface}

\label{sec:demonstration}

In this section, we demonstrate the functionality of our approach through a web-based claim analysis tool that allows users to easily interact with the system and analyze claims within relevant texts, leveraging the backend architecture presented in the previous section. The demo showcases the complete workflow, from entering a scientific passage to obtaining fact-checking results with supporting evidence and explanations. 

The user interface provides a drop-down menu featuring over 30 pre-selected examples from various domains, including biomedicine papers, COVID-related news, social media, and patents (see Figure \ref{fig:screenshot}). These examples were chosen to represent a broad spectrum of platforms and relevant disciplines, allowing users to explore diverse contexts. By selecting any example from the list, the corresponding text will be displayed in the text box below, with a hyperlink to the original source. The text box is fully editable, enabling users to modify the content and analyze their own text.  While the application can process texts up to 10,000 characters, we recommend keeping the input under 2,000 characters for faster results.

As shown in Figure \ref{fig:screenshot}, once the user has selected or entered a text, they can click the \textit{Analyze} button to initiate the claim analysis process. Results are presented as a list of identified claims, each of which can be expanded to reveal detailed information. For each claim, the interface shows:

\textbf{Prediction label:} Whether the claim is supported (green) or refuted (red). Results labeled as \textit{Not Enough Information} by the LLM are not presented to the user.

\textbf{Prediction score:} A normalized probability score that represents the confidence level of the model in its prediction. This score is derived from the statistical outputs of the model, particularly the tokens representing the label string, and it is expressed as a percentage.

\textbf{Evidence:} The related document selected by the retrieval module, along with its DOI. The specific sentence(s) in the abstract of the paper that were most influential for the prediction according to the LLM are highlighted.

\textbf{Rationale:} A justification of the reasoning behind the classification, providing additional insight into the decision-making process carried out by the model.

This interactive interface provides an intuitive and user-friendly environment for testing and exploring claim analysis in various health-related texts. The analysis goes through the entire pipeline, offering a label for the claims extracted along with the retrieved evidence. When a claim receives conflicting labels from different pieces of evidence, our system returns all relevant pairs, allowing users to assess which is more accurate. Since scientific knowledge evolves, evidence considered true at one time may later be contradicted. We recognize the value of providing all supporting and refuting evidence along with their publication dates.

\section{Experimentation and Results}
\label{sec:exp}
We evaluate our system following a three-stage approach. First, we evaluate SciClaims' modules against some baselines using SciFact as benchmark. Then, we evaluate the performance of the system at every step of the pipeline using an LLM as a judge. Finally, we perform a human evaluation for the whole system.

\subsection{Evaluation in SciFact}
\label{sec:scifact_eval}
We use SciFact, a benchmark dataset focused on biomedical literature, to evaluate SciClaims. It includes 1,400 expert-authored scientific claims, each linked to evidence-containing abstracts and annotated with a verification label. 

SciFact supports two complementary evaluations: (1) claim extraction, by comparing system-generated claims to human-written ones using shared source documents, and (2) claim verification, by assessing how accurately the system labels claim-evidence pairs.

\subsubsection{Claim extraction}
To evaluate the claim extraction module, we consider the expert-written claims from SciFact as the ground truth for each source paragraph.

\begin{table}[H]
\centering
\begin{adjustbox}{width=0.48\textwidth}
\begin{tabular}{ lcccc } 
 \hline
 Method & Rouge1 & Rouge2 & RougeL & Similarity Score \\ 
 \hline
 Sentence tokenizer & 0.3313 & \textbf{0.1980} & 0.3030 & 0.7211 \\ 
 \cite{pan-etal-2021-zero} & 0.2780 & 0.1334 & 0.2555 & 0.6561 \\ 
 \cite{wright-etal-2022-generating} & 0.2525 & 0.0762 & 0.2220 & 0.6475 \\ 
 \hline
 Noun Phrases Gen & 0.2567 & 0.0953 & 0.2293 & 0.6693 \\ 
 SciClaims & \textbf{0.3387} & 0.1896 & \textbf{0.3084} & \textbf{0.7250} \\
 \hline
\end{tabular}
\end{adjustbox}
\caption{Claim quality scores in SciFact} 
\label{table:cg_scores}
\end{table}

As a first baseline, we select a sentence tokenizer, treating each sentence in the source paragraph as a claim. Next, we select two baselines \cite{pan-etal-2021-zero, wright-etal-2022-generating} that use a pipeline with two transformer models to generate claims from the source paragraph through entity extraction. The first model generates a question-answer pair, where the answer is the entity, while the second reformulates it as a claim. We also introduce a method where we propose to build the claims around noun phrases rather than named entities.

We evaluate the quality of claim generation by matching each system-generated claim with its most similar gold claim from the same paragraph. We discard pairs with a Levenshtein similarity score below 0.3, a threshold empirically tuned to balance recall and noise. For valid matches, we compute ROUGE-1, ROUGE-2, ROUGE-L, and a semantic similarity score using a DeBERTa model fine-tuned on the STS-B dataset. As shown in Table \ref{table:cg_scores}, SciClaims achieves the highest scores in ROUGE-1, ROUGEL, and semantic similarity, indicating more faithful and informative claim generation.

\subsubsection{Claim verification}
We evaluate SciClaims on SciFact’s test set, comparing its label accuracy for claim-evidence pairs with existing approaches.

\begin{table}[H]
\centering
\begin{adjustbox}{width=0.48\textwidth}
\begin{tabular}{ lccc } 
 \hline
 Architecture & Precision & Recall & F1-Score \\ 
 \hline
 Roberta-base* & 0.4662 & 0.5963 & 0.5220 \\
 MultiVerS* & \textbf{0.7286} & \textbf{0.7321} & \textbf{0.7303} \\ 
 \hline
 SciClaims (LLaMA3 8B Instruct) & 0.7034 & 0.6863 & 0.6788 \\
 SciClaims (Qwen2 7B Instruct) & 0.6649 & 0.6677 & 0.6439\\
 SciClaims (Phi3 Small 8K) & 0.7100 & 0.6894 & 0.6525 \\
 SciClaims (OLMO2 7B Instruct) & 0.6379 & 0.6118 & 0.5886 \\
 \hline
\end{tabular}
\end{adjustbox}
\caption{Precision, recall and F1-Scores in Claim Verification of SciFact test set. RoBERTa-base and MultiVerS are fine-tuned models, while SciClaims is zero-shot. In parenthesis, the backbone used in each SciClaims configuration} 
\label{table:cfc_scores}
\end{table}

 We compare SciClaims with two strong baselines: a RoBERTa-base classifier \cite{liu2019robertarobustlyoptimizedbert} and MultiVerS \cite{wadden2022multiversimprovingscientificclaim}, a longformer-based model for claim verification. Both baselines are fine-tuned on SciFact. While MultiVerS achieves the highest F1-score, SciClaims performs competitively despite operating in a zero-shot setting. As shown in Table \ref{table:cfc_scores}, the relatively narrow performance gap demonstrates SciClaims’ strong generalization ability without task specific training. Furthermore, as shown in Table \ref{table:cfc_scores}, we evaluate SciClaims with different LLMs with similar sizes as backbone, being LLaMA3 8B Instruct the one which offered the best performance.

\subsection{Evaluation with a judge model}
\label{sec:judge}
The second stage of our evaluation provides a comprehensive assessment of the entire system using a LLM as judge. Based on the Judge Arena leaderboard\footnote{https://huggingface.co/spaces/AtlaAI/judge-arena}, we select Qwen 2.5 72B turbo as judge, since it is the first ranked model with open weights and a higher parameter count than our system's LLM (Llama3 8B Instruct). We chose its 4-bit quantization version (Qwen2.5 72B AWQ\footnote{huggingface.co/Qwen/Qwen2.5-72B-Instruct-AWQ}) due to hardware limitations. For the evaluation sample, we randomly select 120 documents from the PubMed dataset presented in section \ref{sec:system}. The evaluation is conducted in three phases: Claim quality, document retrieval, and claim verification and full-system evaluation.

\subsubsection{Claim quality evaluation}

In this phase, we evaluate the quality of the claims generated by SciClaims and compare it to other methods, using the same baselines mentioned in Section \ref{sec:scifact_eval}.

\begin{table*}[ht]
\footnotesize
\centering
\resizebox{\textwidth}{!}{ 
\begin{tabular}{lcccccccccc}
\toprule
 Method & Q1 & Q2 & Q3 & Q4 & Q5 & Q6 & Q7 & Q8 & Correct Claims (\%) & Candidate claims \\ \midrule
Sentence tokenizer & \textbf{0.9987} & 0.9100 & 0.8436 & 0.8709 & 0.9322 & 0.9009 & 0.3703 & 0.9596 & 32.46 & 767 \\
\cite{pan-etal-2021-zero} & 0.2449 & 0.1303 & 0.1618 & 0.1371 & 0.2112 & 0.1169 & 0.0337 & 0.1641 & 3.15 & 445 \\
\cite{wright-etal-2022-generating} & 0.8043 & 0.4714 & 0.5040 & 0.3842 & 0.5593 & 0.4829 & 0.1641 & 0.6249 & 11.28 & 4175 \\
Noun Phrases Gen & 0.3001 & 0.1562 & 0.1857 & 0.1209 & 0.2388 & 0.1320 & 0.0316 & 0.2203 & 2.42 & \textbf{4335} \\\midrule
SciClaims & 0.9949 & 0.9474 & 0.9718 & 0.9089 & 0.9320 & 0.9345 & 0.5238 & 0.9756 & 47.37 & 779 \\
SciClaims+CDP & 0.9942 & 0.9690 & 0.9845 & 0.9380 & 0.9574 & 0.9593 & 0.5930 & \textbf{0.9903} & 55.43 & 516 \\
SciClaims+CDP+CR & 0.9923 & \textbf{0.9787} & \textbf{0.9884} & \textbf{0.9845} & \textbf{0.9806} & \textbf{0.9748} & \textbf{0.7810} & \textbf{0.9903} & \textbf{76.36} & 516 \\ 
\bottomrule
\end{tabular}
}
\caption{Claim extraction (phase 1) evaluation results with a judge model (Qwen2.5 72B AWQ).}
\label{table:phase1_results}
\end{table*}

\begin{table}[ht]
\footnotesize
\centering
\begin{adjustbox}{width=0.48\textwidth}
\begin{tabular}{l ccc ccc}
\toprule
\multirow{2}{*}{Claims} & \multicolumn{3}{c}{All Claims} & \multicolumn{3}{c}{Correct Claims} \\ 
\cmidrule(lr){2-4} \cmidrule(lr){5-7}
 & R@1 & R@3 & R@5 & R@1 & R@3 & R@5 \\ \midrule
Sentence tokenizer & 0.5591 & 0.7205 & 0.7795 & 0.6265 & 0.7390 & 0.7871 \\
\cite{pan-etal-2021-zero} & 0.2970 & 0.4653 & 0.5248 & 0.6429 & 0.7143 & 0.7143 \\
\cite{wright-etal-2022-generating} & \textbf{0.5860} & \textbf{0.7474} & \textbf{0.8038} & \textbf{0.7113} & \textbf{0.8365} & \textbf{0.8726} \\
Noun Phrases Gen & 0.4634 & 0.6456 & 0.7013 & 0.6286 & 0.7619 & 0.8095 \\
\midrule
SciClaims & 0.5045 & 0.6645 & 0.7135 & 0.5257 & 0.6938 & 0.7344 \\
SciClaims+CDP & 0.5146 & 0.6940 & 0.7466 & 0.5944 & 0.7448 & 0.7867 \\
SciClaims+CDP+CR & 0.4727 & 0.6250 & 0.6777 & 0.5102 & 0.6675 & 0.7107 \\
\bottomrule
\end{tabular}
\end{adjustbox}
\caption{Document retrieval (phase 2) evaluation results  with a judge model (Qwen2.5 72B AWQ).}
\label{table:phase2_results}
\end{table}

We devised a questionnaire consisting of eight yes/no questions (see Appendix \ref{sec:questions}) to capture the desired properties in a correct claim and ask the judge model to answer it. The first question requires context from the source paragraph, while the remaining questions focus solely on the claim. Correct claims need to receive a \textit{Yes} to all questions. Table \ref{table:phase1_results} shows that our LLM-based system outperforms all other methods, scoring 15 points higher than the second-best, the sentence tokenizer.

\begin{table*}[ht]
\footnotesize
\centering
\resizebox{\textwidth}{!}{ 
\begin{tabular}{ll | cc cc | c | c}
\toprule
\multicolumn{2}{c}{System} & \multicolumn{2}{c}{All Claims} & \multicolumn{2}{c}{Correct Claims} \\ 
Extraction Module & Verification Module & Label Accuracy & Not NEI (\%) & Label Accuracy & Not NEI (\%) & System Score & Time/doc (secs.) \\ \midrule
Sentence tokenizer & MultiVerS & 0.5494 & 4.77 & 0.5127 & 4.82 & 0.1664 & 2.36 \\
\cite{pan-etal-2021-zero} & MultiVerS & \textbf{0.7591} & 2.31 & 0.5714 & 11.90 & 0.0180 & 2.17\\
\cite{wright-etal-2022-generating} & MultiVerS & 0.5015 & 5.53 & 0.4268 & 8.35 & 0.0482 & 14.25 \\
Noun Phrases Gen & MultiVerS & 0.5965 & 3.08 & 0.4825 & 6.67 & 0.0117 & 15.01 \\
\midrule
SciClaims & MultiVerS & 0.5944 & 6.02 & 0.5709 & 6.50 & 0.2704 & 2.86 \\
SciClaims+CDP & MultiVerS & 0.5867 & 5.71 & 0.5361 & 5.60 & 0.2971 & 1.82 \\
SciClaims+CDP+CR & MultiVerS & 0.6204 & 3.72 & 0.5922 & 7.44 & 0.4522 & 2.93 \\
\hline
Sentence tokenizer & SciClaims & 0.6041 & 61.20 & 0.6693 & 59.17 & 0.2173 & 11.72 \\
\cite{pan-etal-2021-zero} & SciClaims & 0.6568 & 39.60 & 0.619 & 71.43 & 0.0195 & 7.61 \\
\cite{wright-etal-2022-generating} & SciClaims & 0.6397 & \textbf{69.44} & \textbf{0.7098} & \textbf{74.66} & 0.0801 & 65.28\\
Noun Phrases Gen & SciClaims & 0.6608 & 53.02 & 0.6762 & 70.48 & 0.0164 & 68.01\\
\midrule
SciClaims & SciClaims & 0.6361 & 58.49 & 0.6567 & 59.71 & 0.3111 & 12.38 \\
SciClaims+CDP & SciClaims & 0.6296 & 58.87 & 0.6364 & 62.47 & 0.3527 & 8.13 \\
SciClaims+CDP+CR & SciClaims & 0.6589 & 53.06 & 0.6574 & 54.23 & \textbf{0.5020} & 9.24 \\
\bottomrule
\end{tabular}
}
\caption{Verification evaluation (phase 3) results with a judge model (Qwen2.5 72B AWQ). The table shows the label accuracy of each combination of extraction and verification module. The Correct Claims column counts only those considered correct in the Phase 1 questionnaire. Not NEI represents the proportion of claim–evidence pairs labeled as SUPPORT or REFUTE, rather than NEI (Not Enough Information). The last two columns summarize system-level performance: Processing Time per Document is the average time (in seconds) each system takes to process a single document, and System Score reflects the ratio of correctly labeled and well-formed claims across the system.}
\label{table:phase3_results}
\end{table*}

To further enhance the results, we refined our claim extraction prompts with two optimizations:

\textbf{Claim Definition Properties (CDP)}. Here we enhance the prompt by incorporating the characteristics of a claim, such as precision, conciseness, or check-worthiness. These characteristics are derived from the questionnaire used to evaluate the quality of the claims. The goal is to guide the LLM to adhere to these criteria when generating the list of claims.

\textbf{Claim Refinement (CR)}: This upgrade involves a follow-up call to the LLM in order to refine the initial list of claims. For each candidate claim, we pair it with the source paragraph and ask the LLM to refine the claim based on the same criteria (precision, conciseness, check-worthiness, etc.). This step aims to eliminate poorly formed claims and reinforce the application of the specified criteria.

These two upgrades result in an additional 29-point increase in the percentage of correct claims generated by SciClaims while reducing the number of candidate claims generated by our approach. Notably, QA-oriented baselines, such as \cite{wright-etal-2022-generating} and our noun phrase-based generation method, generate significantly larger quantities of claims compared to the other methods in Table \ref{table:phase1_results}.

\subsubsection{Document retrieval evaluation}

We assess document relevance by asking the judge model whether each retrieved paragraph aids claim verification. Recall is evaluated at k = {1,3,5} retrieved documents per claim. As shown in Table \ref{table:phase2_results}, claims generated by \cite{wright-etal-2022-generating} retrieve more potentially relevant documents than SciClaims. However, SciClaims performs well, retrieving at least one relevant document in 75\% of cases when fetching five documents per claim, a common real-world scenario. This highlights its balance between claim accuracy and document relevance. QA-oriented methods generate more compact, less specific claims than LLMs, increasing the likelihood of retrieving related documents.

\subsubsection{Claim verification and full-system}

The verification task involves predicting the veracity of claim-evidence pairs, classifying each as \textit{SUPPORT}, \textit{REFUTE}, or \textit{NOT ENOUGH INFORMATION (NEI)}. Thus we evaluate our system's label accuracy by asking to the judge model whether the assigned label is correct. We compare SciClaims results
with MultiVerS \cite{wadden2022multiversimprovingscientificclaim}.

SciClaims demonstrates superior claim verification accuracy, outperforming the fine-tuned MultiVerS \cite{wadden2022multiversimprovingscientificclaim} model despite operating in a zero-shot setting. As shown in Table \ref{table:phase3_results}, SciClaims consistently produces more accurate labels across all claim generation methods, except \cite{pan-etal-2021-zero}. Notably, SciClaims returns significantly fewer \textit{NEI} labels than MultiVerS, providing greater value to users by delivering more definitive \textit{SUPPORT} or \textit{REFUTE} labels. The unusually high accuracy of \cite{pan-etal-2021-zero}. likely stems from its excessive \textit{NEI} labels, which simplify label selection. Furthermore, the SciClaims+CDP+CR system achieves the highest overall performance, correctly labeling 50\% of generated claims, outperforming the next-best MultiVerS-based system by over five points.

Table \ref{table:phase3_results} also reports the average processing time per document for each system configuration. Systems using MultiVerS as the verification module tend to be the fastest overall, with the exception of those paired with claim extraction modules like \cite{wright-etal-2022-generating} and Noun Phrases Generation, which produce a high number of claims and thus increase runtime. Among all configurations, those using SciClaims+CDP+CR for claim generation—paired with either SciClaims or MultiVerS for verification—offer the best trade-off between predictive accuracy and time efficiency. Importantly, only the configuration that uses SciClaims as both the generation and verification module can run on a single 24GB GPU, making it uniquely suitable for real-world deployment.

\subsection{Human evaluation}
To complement our automated evaluation with LLMs, we conducted a human evaluation to assess the quality, relevance, and usability of SciClaims outputs. Five NLP-experts independently reviewed a sample of 154 PubMed documents processed by the system. The annotation tasks were divided into three phases, following the same structure as our LLM-based evaluation (see Section~\ref{sec:judge}).

\textbf{Claim quality evaluation.} Annotators were shown an input paragraph and one randomly selected claim extracted by SciClaims. They answered the same eight binary questions used in the LLM-based evaluation (see Appendix~\ref{sec:questions}), assessing conciseness, precision, and other key dimensions. Table \ref{table:results_human} shows that human judgments closely align with the judge model, with 70.5\% of the claims meeting all eight criteria.

 \begin{table}[ht]
\footnotesize
\centering
\begin{adjustbox}{width=0.48\textwidth}
\begin{tabular}{lc | lc}
\toprule
Claim Quality & Score & Rationale Quality & Score \\ \midrule
Q1 (grounding) & 0.9048 \\
Q2 (grammar) & 0.9810 \\
Q3 (completeness) & 0.9810 & RQ1 (justification) & 0.8571\\
Q4 (precision) & 0.8667 & RQ2 (relevance) & 0.9206 \\
Q5 (relevance) & 0.8952 & RQ3 (completeness) & 0.7937\\
Q6 (conciseness) & 0.8571 \\
Q7 (self-contained) & 0.9238\\
Q8 (contribution) & 0.8857 \\\midrule
Correct Claims (\%) & \textbf{70.47} & Correct Rationales (\%) & \textbf{74.61}\\
\bottomrule
\end{tabular}
\end{adjustbox}
\caption{Claim and rationale evaluation of SciClaims by human annotators. Score indicates the overall ratio of 'yes' responses to the question along the annotators.}
\label{table:results_human}
\end{table}

\textbf{Document retrieval evaluation.} Annotators were presented with a claim and its corresponding retrieved paragraph from SciClaims. They were asked whether the retrieved evidence provided sufficient information to verify the claim. In 60\% of cases, annotators judged the paragraph as informative enough for claim verification, indicating moderate effectiveness of the retrieval module in real-world scenarios.

\textbf{Claim verification evaluation.} In this phase, annotators were shown a claim, its evidence, the system’s predicted label, and the corresponding rationale. They rated the label’s accuracy on a five-point scale, from 1 (completely inaccurate) to 5 (highly accurate). SciClaims achieved a strong average score of 4.40, reflecting high alignment between system predictions and human judgment. Additionally, annotators evaluate the quality of the generated rationale using three yes/no questions: (RQ1) Is the label justified by the rationale? (RQ2) Does the rationale focus on relevant information? (RQ3) Does the rationale provide enough context to understand the label?. As shown in Table \ref{table:results_human}, the responses from annotators indicate high satisfaction with the relevance and justification of rationales, though they also noted that completeness could be improved.

\section{Conclusion}
We presented SciClaims, a practical, end-to-end system for scientific claim analysis in the biomedical domain. Through an interactive web-based interface, users can extract, verify, and explore scientific claims with evidence-backed rationales, all powered by LLMs and a curated biomedical corpus. SciClaims is designed for usability and speed, and it is already being explored in real-world settings such as pharmaceutical patent analysis and systematic literature reviews. By combining explainable outputs with efficient infrastructure, it provides a robust tool for researchers, clinicians, and analysts who need to validate scientific information quickly and reliably. The system is openly accessible and ready for demonstration, offering an engaging experience to showcase the capabilities of modern LLM-based scientific reasoning.


\clearpage
\section*{Acknowledgments}
The authors gratefully acknowledge the EU Large Language Models for EU (LLMs4EU) project (DIGITAL-20234-AI-06-LANGUAGE-01) and the HORIZON FAIR to Adapt to Climate Change (FAIR2Adapt) grant (agreement 101188256) for their support during the development of this work.   


\bibliography{custom}

\clearpage
\appendix

\section{Prompts}
\label{sec:prompts}

\subsection{SciClaims claim extraction first step}

\begin{lstlisting}[breaklines=true, columns=flexible, basicstyle=\footnotesize]
<|begin_of_text|><|start_header_id|>system<|end_header_id|>
Your task is to generate a list with the main factual claims stated in a text. A factual claim makes an assertion about something regarding the subject matter that can be proved or contradicted with factual evidence. Factual claims must be expressed as meaningful, self-contained sentences. Do not include narrative context and disregard absolutely ALL self-referential parts.

Arrange your output using the format:
-- claim
-- claim
-- claim.

<|begin_of_text|><|start_header_id|>user<|end_header_id|>
TEXT: {text}
\end{lstlisting}

\subsection{SciClaims+CDP}
\begin{lstlisting}[breaklines=true, columns=flexible, basicstyle=\footnotesize]
<|begin_of_text|><|start_header_id|>system<|end_header_id|>
Identify and list the main scientific claims stated in a passage. Each claim must satisfy the following criteria:
- **Convey an insight, interpretation, or conclusion drawn from the passage that is testable and generalizable**: The claim should assert an outcome, capability, or effect rather than merely describing a method, aim, or process.
    - Example (Good): "Neural networks outperform decision trees in image classification tasks." (Testable outcome)
    - Example (Bad): "The proposed method aims to use neural networks for better image classification." (Descriptive, not assertive)
- **Be expressed as a meaningful, self-contained statement**: Each claim should be fully understandable on its own, without needing context from the passage or other claims. It must convey a complete, independent idea. If referencing a study, survey, result, or process, phrase it as a general, verifiable claim.
    - Example (Good): "The Amazon rainforest is home to over 10 million species."
    - Example (Bad): "As mentioned earlier, the Amazon is one of the most biodiverse places in the world." (Requires prior context and doesn't stand alone.)
- **Emphasize generalization and scientific assertion**: Avoid descriptive or narrative conclusions.
    - Example (Good): "Exposure to blue light before sleep can reduce melatonin production." (Generalized, testable assertion)
    - Example (Bad): "The study investigates how exposure to blue light before sleep affects melatonin production." (Descriptive of method, not a general claim)
- **Be clear and concise**: Use straightforward language without unnecessary words.
    - Example (Good): "The Eiffel Tower is in Paris."
    - Example (Bad): "The Eiffel Tower, which is one of the most iconic landmarks in Europe and attracts millions of tourists every year, is located in Paris, France."
- **Exclude narrative context**: Focus on the factual assertion itself, not the surrounding story or background information.
    - Example (Good): "Water boils at 100C under normal atmospheric pressure."
    - Example (Bad): "In many cultures, people have believed for centuries that water boils at 100C, as scientists confirmed in the 18th century." (Includes unnecessary background information.)
- **Disregard all self-referential content**: Ignore any statements referring to the passage itself or the author intentions.
    - Example (Good): "The Earth orbits the Sun."
    - Example (Bad): "The study explains how the Earth orbits the Sun." (This is self-referential and refers to the passage itself.)
- **Be precise and objective**: Avoid ambiguity, subjective interpretation, or vague statements. Present claims as clear, verifiable facts.
    - Example (Good): "The Great Wall of China stretches approximately 13,000 miles."
    - Example (Bad): "The Great Wall of China is pretty long." (Vague and subjective.)
- **Be relevant to the broader debate or public discourse**: Focus on verification-worthy claims that introduce new information rather than merely restating common knowledge.
    - Example (Good): "The global temperature has increased by about 1C since the late 19th century."
    - Example (Bad): "The Earth is a planet." (Common knowledge and not contributing new, verifiable information.)

Present the output using the following format:
-- claim
-- claim
-- claim

<|begin_of_text|><|start_header_id|>user<|end_header_id|>
PASSAGE: {text}
\end{lstlisting}

\subsection{SciClaims claim extraction second step (refinement)}
\begin{lstlisting}[breaklines=true, columns=flexible, basicstyle=\footnotesize]
Now, using the information given by the passage, reformulate each individual claim to be fully understandable by itself, even without having the context from the passage or the rest of the claims from the list. Change the terminology or add context information to each claim if necessary.
\end{lstlisting}

\subsection{SciClaims+CDP+CR (refinement)}
\begin{lstlisting}[breaklines=true, columns=flexible, basicstyle=\footnotesize]
<|begin_of_text|><|start_header_id|>system<|end_header_id|>
Given a claim and the passage it was extracted from, reformulate the claim to fully adhere to **ALL** the following criteria.
- **Convey an insight, interpretation, or conclusion drawn from the passage that is testable and generalizable**: The claim should assert an outcome, capability, or effect rather than merely describing a method, aim, or process.
    - Example (Good): "Neural networks outperform decision trees in image classification tasks." (Testable outcome)
    - Example (Bad): "The proposed method aims to use neural networks for better image classification." (Descriptive, not assertive)
- **Be expressed as a meaningful, self-contained statement**: Each claim should be fully understandable on its own, without needing context from the passage or other claims. It must convey a complete, independent idea. If referencing a study, survey, result, or process, phrase it as a general, verifiable claim.
    - Example (Good): "The Amazon rainforest is home to over 10 million species."
    - Example (Bad): "As mentioned earlier, the Amazon is one of the most biodiverse places in the world." (Requires prior context and doesn't stand alone.)
- **Emphasize generalization and scientific assertion**: Avoid descriptive or narrative conclusions.
    - Example (Good): "Exposure to blue light before sleep can reduce melatonin production." (Generalized, testable assertion)
    - Example (Bad): "The study investigates how exposure to blue light before sleep affects melatonin production." (Descriptive of method, not a general claim)
- **Be clear and concise**: Use straightforward language without unnecessary words.
    - Example (Good): "The Eiffel Tower is in Paris."
    - Example (Bad): "The Eiffel Tower, which is one of the most iconic landmarks in Europe and attracts millions of tourists every year, is located in Paris, France."
- **Exclude narrative context**: Focus on the factual assertion itself, not the surrounding story or background information.
    - Example (Good): "Water boils at 100C under normal atmospheric pressure."
    - Example (Bad): "In many cultures, people have believed for centuries that water boils at 100C, as scientists confirmed in the 18th century." (Includes unnecessary background information.)
- **Disregard all self-referential content**: Ignore any statements referring to the passage itself or the author intentions.
    - Example (Good): "The Earth orbits the Sun."
    - Example (Bad): "The study explains how the Earth orbits the Sun." (This is self-referential and refers to the passage itself.)
- **Be precise and objective**: Avoid ambiguity, subjective interpretation, or vague statements. Present claims as clear, verifiable facts.
    - Example (Good): "The Great Wall of China stretches approximately 13,000 miles."
    - Example (Bad): "The Great Wall of China is pretty long." (Vague and subjective.)
- **Be relevant to the broader debate or public discourse**: Focus on verification-worthy claims that introduce new information rather than merely restating common knowledge.
    - Example (Good): "The global temperature has increased by about 1C since the late 19th century."
    - Example (Bad): "The Earth is a planet." (Common knowledge and not contributing new, verifiable information.)

Present the output using the following format:
{"original_claim": <str>, "refined_claim": <str>, "rationale": <str>}

<|begin_of_text|><|start_header_id|>user<|end_header_id|>
CLAIM: {claim}
PASSAGE: {text}
\end{lstlisting}

\subsection{SciClaims claim verification}
\begin{lstlisting}[breaklines=true, columns=flexible, basicstyle=\footnotesize]
<|begin_of_text|><|start_header_id|>system<|end_header_id|>
You are a claim analyst. Upon receiving a claim and an evidence, your task is to figure out if the claim is either supported, contradicted or unrelated based exclusively on the evidence.

- If you are confident that the claim is supported by the evidence your answer will be "SUPPORT".
- If you are certain that the evidence directly contradicts the claim, your answer will be "CONTRADICT". Please note that if the claim is just not mentioned in the evidence, or if it is unrelated to the evidence, it does not mean it is contradicted. For that cases, the answer will be "NEI".
- If the evidence does not contain enough information or if it is not related to the claim, your answer will be "NEI", which stands for Not Enough Information.

Arrange the output as a JSON dictionary with the keys "response" and "evidence" and ensure your output is JSON-valid.
- The "response" values can only be "SUPPORT", "CONTRADICT" or "NEI".
- The "evidence" value must be a list of sentences from the evidence which are more related to your decision. If the decision is "NEI", this field will be empty.

<|begin_of_text|><|start_header_id|>user<|end_header_id|>
CLAIM: {claim}
EVIDENCE: {evidence}
\end{lstlisting}

\subsection{Phase 1 evaluation with judge model (Q1)}
\begin{lstlisting}[breaklines=true, columns=flexible, basicstyle=\footnotesize]
<|begin_of_text|><|start_header_id|>system<|end_header_id|>
Given a sentence and a paragraph, answer the following question. Use exclusively the content of the paragraph to answer the question. 

Is the sentence supported by the paragraph?

Return your response as a json dictionary, following this structure: {"answer":<Yes/No>, "rationale": <str>}

<|begin_of_text|><|start_header_id|>user<|end_header_id|>
SENTENCE: {claim}
PARAGRAPH: {text}
\end{lstlisting}

\subsection{Phase 1 evaluation with judge model (Q2-8)}
\begin{lstlisting}[breaklines=true, columns=flexible, basicstyle=\footnotesize]
Given a claim, answer the following question. 

{QUESTION}

Return your response as a json dictionary, following this structure: {"answer":<Yes/No>, "rationale": <str>}

<|begin_of_text|><|start_header_id|>user<|end_header_id|>
CLAIM: {claim}

\end{lstlisting}

\subsection{Phase 2 evaluation with judge model}
\begin{lstlisting}[breaklines=true, columns=flexible, basicstyle=\footnotesize]
<|begin_of_text|><|start_header_id|>system<|end_header_id|>
Given a claim and a paragraph, answer the following question. 

Is the information contained in the paragraph useful to verify the claim?

Return your response as a json dictionary, following this structure: {"answer":<Yes/No>, "rationale": <str>}

<|begin_of_text|><|start_header_id|>user<|end_header_id|>
CLAIM: {claim}
PARAGRAPH: {text}
\end{lstlisting}

\subsection{Phase 3 evaluation with judge model}
\begin{lstlisting}[breaklines=true, columns=flexible, basicstyle=\footnotesize]
<|begin_of_text|><|start_header_id|>system<|end_header_id|>
Given a claim and a paragraph, answer the following question. 

Is the claim {SUPPORTED/REFUTED} by the paragraph?

Return your response as a json dictionary, following this structure: {"answer":<Yes/No>, "rationale": <str>}

<|begin_of_text|><|start_header_id|>user<|end_header_id|>
CLAIM: {claim}
PARAGRAPH: {text}
\end{lstlisting}

\clearpage
\section{Claim Quality Questionnaire}
\label{sec:questions}
\begin{table}[ht]
\footnotesize
\centering
\begin{tabular}{p{0.04\linewidth} | p{0.81\linewidth}} 
 \hline
 \textbf{Id} & \textbf{Question} \\ 
 \hline
 Q1 & Is the claim grounded by the original text? \\
 Q2 & Is the claim grammatically correct? \\ 
 Q3 & Does the claim have all the necessary components (subject, predicate, and relevant qualifiers) to form a complete thought? \\
 Q4 & Is the claim precise and specific rather than vague? \\ 
 Q5 & Does the claim introduce new information rather than just restating common knowledge? \\
 Q6 & Is the claim concise without losing essential information? \\ 
 Q7 & Does the claim provide enough information to be understood independently? \\
 Q8 & Would verifying the claim add value to public knowledge? \\ 
 \hline
\end{tabular}
\caption{Questions asked to the LLM judge and human annotators to evaluate the quality of the generated claims.}
\label{table:questions}
\end{table}
\end{document}